\documentclass[11pt]{article}

\usepackage[final]{acl}

\usepackage{times}
\usepackage{latexsym}

\usepackage[T1]{fontenc}
\usepackage[utf8]{inputenc}

\usepackage{microtype}
\usepackage{inconsolata}

\usepackage{graphicx}
\usepackage{amsmath}
\usepackage{mathtools}
\usepackage{amssymb}
\usepackage{algorithm}
\usepackage{algorithmic}
\usepackage{enumitem}
\usepackage{booktabs}
\usepackage{stfloats}
\usepackage{cuted}
\usepackage{array,tabularx,booktabs,multirow}
\usepackage{adjustbox}
\usepackage{float}
\usepackage{colortbl}
\definecolor{OursRow}{HTML}{EAF1FB}
\definecolor{ImpGreen}{HTML}{1B7F3F}
\newcommand{\imp}[1]{{\color{ImpGreen}\scriptsize\,($\Delta$#1)}}

\usepackage{array}
\usepackage{multirow}

\newcolumntype{L}[1]{>{\raggedright\let\newline\\\arraybackslash\hspace{0pt}}p{#1}}

\title{PhoenixNest-Video: Evidence-Grounded Multimodal Agent Framework for Automated Video Interview Assessment}
\author{Yuxuan Fan \quad Miaojun Huang \quad Haimei Zhang \quad Jingshen Wu \quad Hao Liu\\
  The Hong Kong University of Science and Technology (Guangzhou)}

\begin{document}
\maketitle

\begin{abstract}
Interview assessment requires per-criterion judgments grounded in behavioral evidence, yet surging applicant volumes have made human-only evaluation costly and inconsistent, while existing AI approaches yield opaque scores without traceable rationale. We introduce PhoenixNest-Video, an evidence-grounded multimodal agent framework for automated video interview assessment. It builds a semantic video graph as structured working memory, performs rubric-conditioned retrieval with cross-modal verification across visual, audio, and textual streams, and produces per-criterion scores anchored to the candidate's materials. A Scorer trained via Rubrics-based Reinforcement Learning with dual rewards for rubric alignment and score-level differentiation internalizes the discriminative structure of multi-level rubrics. PhoenixNest-Video attains 91.50\% grade-level accuracy on VInterview-2025, outperforming substantially larger proprietary models. A compact, rubric-grounded agent therefore scores candidates in closer agreement with an expert panel than direct prompting of much larger models, and exposes the evidence behind each score for human review.
\end{abstract}

\section{Introduction}
\label{sec:intro}
Interviews have long served as a core step in selecting candidates across graduate admissions and professional hiring~\cite{wiens1976assessment, franklin2024work, albaroudi2024comprehensive}. To reach fair and defensible decisions, institutions require evaluators to score each candidate along multiple criteria, justify every score, and cite specific moments and materials from the interview as supporting evidence~\cite{maude2022holistic}. Yet application volumes have surged sharply in recent years, with Common Application submissions rising over 37\% between 2021 and 2025 to more than 7.6 million~\cite{magouirk2025deadline, altbach2019trends}, placing unprecedented pressure on this evidence-anchored protocol. Sustaining such rigorous per-candidate assessment with human experts alone has become increasingly impractical at this scale. On one hand, recruiting, training, and compensating qualified interviewers across thousands of applicants imposes a substantial labor cost, and expert staffing has not kept pace with demand. On the other hand, even when sufficient raters are available, prolonged large-scale evaluation amplifies well-documented human limitations such as fatigue, anchoring biases, and inconsistent criterion weighting~\cite{conway1995meta, campion1997review, landy1980performance}, eroding the consistency that high-stakes decisions depend on. An automated assistant addresses this directly: it applies one rubric uniformly across a large candidate pool and makes the evidence behind each score inspectable.

\begin{figure}[t]
    \centering
    \includegraphics[width=0.5\textwidth]{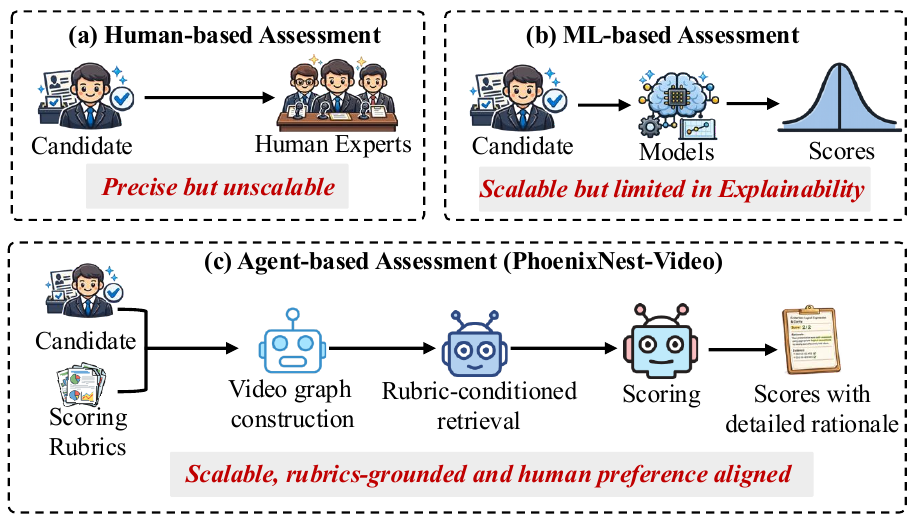}
    \caption{Three paradigms of interview assessment. (a) Human-based assessment is precise but unscalable. (b) ML-based assessment is scalable but opaque. (c) PhoenixNest-Video produces per-criterion scores with verifiable evidence, achieving scalable and evidence-traceable assessment.}
    \label{fig:evolution}
\end{figure}

As illustrated in Figure~\ref{fig:evolution}, interview assessment has evolved through three paradigms. Human panels remain the gold standard in accuracy but cannot scale to current applicant volumes. ML-based approaches~\cite{Naim2015AutomatedAA, Subramaniam2016BimodalFI, Hemamou2019HireNetAH, Agrawal2020LeveragingMB} train supervised models on interview data and gain scalability, yet they yield opaque numeric scores without interpretable rationale, limiting their utility in high-stakes decisions. Multimodal Large Language Models (MLLMs) offer a more capable foundation by processing visual, audio, and textual streams within a unified architecture~\cite{maaz2024video, li2024eald, cheng2024videollama, zhang2025videollama}, yet directly applying them to interview assessment exposes structural challenges that standard training leaves unresolved.

\begin{figure}[t]
    \centering
    \includegraphics[width=0.38\textwidth]{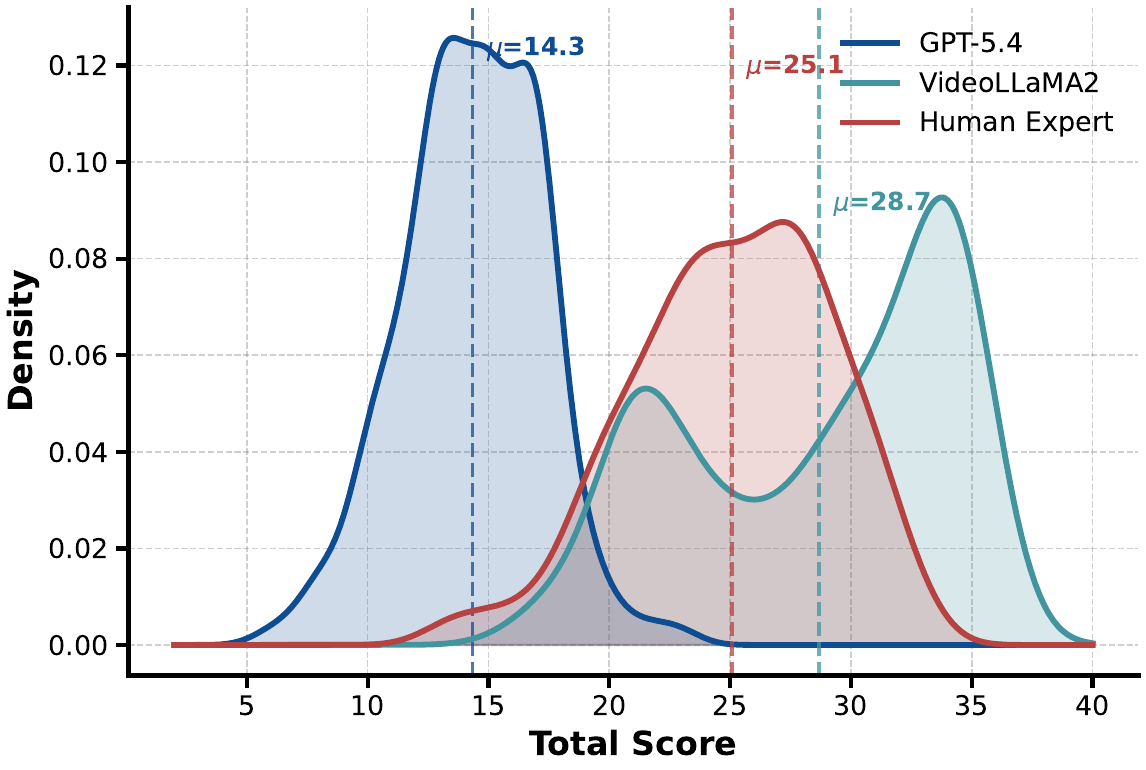}
    \caption{Score distributions of general-purpose MLLMs versus human experts on VInterview-2025. GPT-5.4 systematically underscores within a narrow band, while VideoLLaMA2 overscores with a bimodal pattern. These opposing deviations from the human expert distribution expose complementary failures in calibration and discrimination that motivate rubric-grounded reinforcement learning.}
    \label{fig:challenges}
\end{figure}

Applying MLLMs to criterion-level evaluative judgment exposes three fundamental limitations. First, rubric scoring requires decomposing parallel visual, vocal, and verbal streams and attending to different signals per criterion~\cite{Kim2023FairnessAwareML, Arakawa2022AIFH, Takeuchi2021InitialAO}, yet MLLMs treat video as an undifferentiated stream and cannot organize criterion-relevant evidence across modalities and time~\cite{Wingate2024WhatAI}. Second, as Figure~\ref{fig:challenges} shows, general-purpose MLLMs exhibit systematic scoring biases, some compressing into a narrow low band and others skewing high; optimized for fluent generation and safety-aligned neutrality, they fail to differentiate candidates along rubric-defined dimensions. Third, defensible assessment requires every judgment to be traceable to specific behavioral observations, yet MLLMs produce assessment text without such anchoring and lack the temporal memory to cross-verify what a candidate said, showed, and expressed, leaving their outputs unauditable~\cite{fabeyo2025explainable}. Fine-grained visual distinctions can also trigger hallucinations~\cite{bai2026hallucination}, while visual facts that are neither retained nor verbalized may become inaccessible in later interactions~\cite{chen2026seen}.

To address these challenges, we propose PhoenixNest-Video, an evidence-grounded multimodal agent framework for automated video interview assessment. PhoenixNest-Video constructs a semantic video graph as structured working memory, performs rubric-conditioned retrieval with cross-modal verification to locate and validate criterion-relevant evidence, and produces per-criterion scores anchored to the candidate's materials. A Rubrics-based Reinforcement Learning procedure with dual reward signals for rubric alignment and score-level differentiation trains the scoring module to internalize the discriminative structure of multi-level rubrics. Built on a Qwen3-VL-8B backbone, PhoenixNest-Video achieves 91.50\% grade-level accuracy on VInterview-2025 with the lowest total-score MAE and Wasserstein distance among all baselines, outperforming substantially larger proprietary models, and after retraining on RecruitView~\cite{gupta2025recruitview} attains the best rank-correlation and concordance scores macro-averaged over its 12 regression targets.

The contributions of this paper are summarized as follows:
\begin{itemize}[noitemsep]
    \item We propose PhoenixNest-Video, an evidence-grounded multimodal agent framework that produces criterion-level scores with verifiable video evidence, enabling transparent and auditable automated interview assessment.
    \item We introduce Rubrics-based Reinforcement Learning with dual reward signals for rubric alignment and score-level differentiation, enabling MLLMs to overcome scoring biases and produce rubric-faithful interview assessments.
    \item We evaluate PhoenixNest-Video on VInterview-2025 and RecruitView, where it achieves 91.50\% grade-level accuracy and produces score distributions closely aligned with human experts, demonstrating the viability of compact, evidence-grounded multimodal agents for automated interview assessment.
\end{itemize}

\section{Related Work}
\label{sec:related}
\textbf{Automated Video Interview Assessment.} Automated interview assessment has progressed from handcrafted prosodic and facial features on mock interviews~\cite{Naim2015AutomatedAA, chen2016automatic, chen2017automated} through bimodal personality prediction on short clips~\cite{Subramaniam2016BimodalFI, escalante2018explaining} to hierarchical neural models for asynchronous screening~\cite{Hemamou2019HireNetAH, Agrawal2020LeveragingMB, singhania2020grading}. Recent work has broadened to audio-visual personality benchmarks~\cite{liao2024open}, pose-based analysis~\cite{tang2025pose}, multimodal performance assessment~\cite{li2025listening, inam2026ivas}, naturalistic interview datasets~\cite{gupta2025recruitview}, fairness audits~\cite{mujtaba2025behind, leong2019trust, putra2024mag}, psychometric validation~\cite{liff2024psychometric}, and multi-agent evaluation frameworks~\cite{sun2026comai}. These efforts predominantly output holistic or trait-level scores without rubric grounding or evidence traceability. PhoenixNest-Video addresses this gap by conditioning every score on rubric descriptors and anchoring it to verified candidates' materials.

\noindent \textbf{MLLMs for Video Understanding.} Multimodal Large Language Models (MLLMs) typically bridge a vision encoder with an LLM via instruction tuning~\cite{maaz2024video, zhang2023video}, with subsequent work improving temporal modeling~\cite{cheng2024videollama, zhang2025videollama, shao2025eventvad} and long-form processing through time-aware querying~\cite{ren2024timechat}, dual-rate visual streams~\cite{xu2024slowfast}, and chunk-level compression~\cite{shu2025video}. Reasoning-oriented post-training via reinforcement learning has further improved spatio-temporal reasoning~\cite{feng2025video, li2025videochat, tao2025moss}. Outside video, domain-specific MLLMs have used multidimensional reasoning rewards and iterative visual tools to improve structured reasoning~\cite{hao2026oralgpt, fan2026oralgpt}, while multimodal agents combine tool use with knowledge-grounded retrieval~\cite{hao2026oralagent}. Recent benchmarks have also begun to test what MLLMs miss beyond surface recognition, including intent-level audiovisual understanding~\cite{fan2026musebench}, cross-modal ambiguity resolution~\cite{wang2025mucar}, and competence in humanities and social-science domains where judgment rather than fact retrieval is at stake~\cite{kang2026hssbench}. Multi-agent orchestration provides another route to structured reasoning, both over video~\cite{kugo2025videomultiagents} and in high-stakes decision domains such as legal judgment prediction~\cite{kang2026multimodal}, though benchmarks such as Neptune~\cite{nagrani2024neptune} show that long-horizon temporal reasoning remains a bottleneck. In contrast to these general-purpose systems, PhoenixNest-Video targets evidence-grounded assessment and traceable reasoning over structured interview evaluations.

\section{Task Formulation}
\label{sec:task}

Video interview assessment is widely adopted across educational, professional, and organizational contexts to evaluate candidates through structured multimodal interactions. In a typical protocol, a candidate delivers a presentation before a panel, sometimes followed by a question-and-answer session. Each panelist then independently rates the candidate across a predefined set of criteria and submits a separate score vector, and a final outcome is derived by aggregating the individual scores.

Our objective is to develop an evaluation system that emulates this multi-criterion assessment process. We define the evaluation function $\mathcal{F}$ that maps a candidate's video $V$, optional supplementary materials $M$, a set of $N$ criteria $C = \{c_1, \dots, c_N\}$, and their rubrics $R(c, s)$ providing a textual descriptor for every score level $s$ in an ordinal scale, to per-criterion scores $\mathcal{S} = \{s_1, \dots, s_N\}$, evidence references $E = \{(m_i, \text{rationale}_i)\}$ that link each judgment to specific elements of the candidate's materials, and textual feedback $F$
\begin{equation}
    \mathcal{S}, E, F \gets \mathcal{F}(V, M, C, R).
\end{equation}

This formulation is general. The number of criteria $N$, the score scale, and the rubric content $R$ are parameters supplied by the application setting. We evaluate on two benchmarks with different configurations, as described in Section~\ref{sec:exp}.

\section{Methodology}
\label{sec:method}
\begin{figure*}[t]
    \centering
    \includegraphics[width=\textwidth]{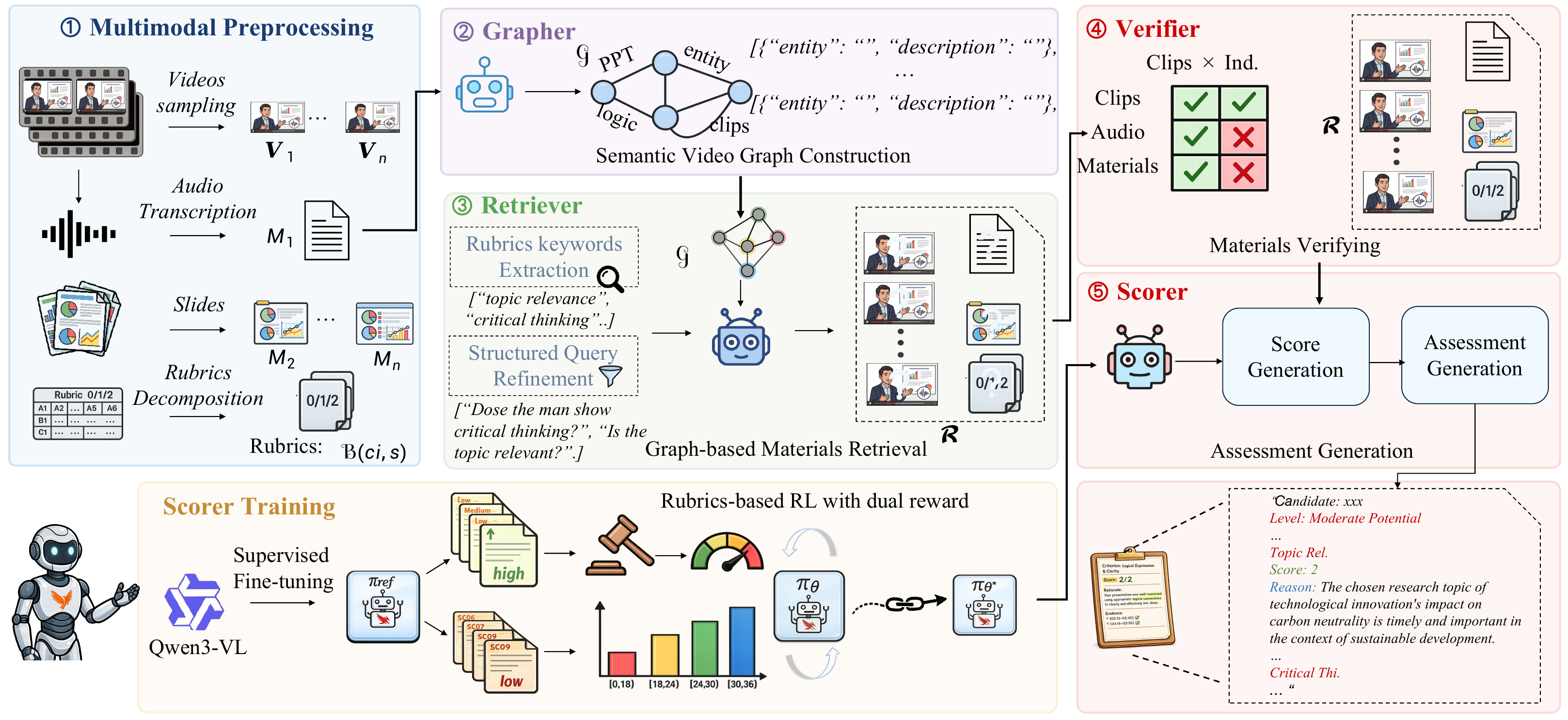}
    \caption{Overview of the PhoenixNest-Video framework. Multimodal Preprocessing converts the interview into aligned visual, audio/transcript, and slide streams, while Rubric Decomposition expands each score descriptor into behavioral indicators $\mathcal{B}(c,s)$. The Grapher constructs a clip-level semantic index; the Retriever selects criterion-relevant candidate clips; the Verifier checks the candidates across modalities; and the trained Scorer $\pi_\theta^*$ produces criterion-level scores, evidence references, and feedback. The Scorer is optimized by supervised fine-tuning followed by rubrics-based reinforcement learning with alignment ($R_{\text{align}}$) and differentiation ($R_{\text{diff}}$) rewards.}
    \label{fig:framework}
\end{figure*}

Figure~\ref{fig:framework} shows the overall architecture. Given a raw interview video and a rubric, PhoenixNest-Video first runs multimodal preprocessing (Section~\ref{sec:preprocessing}) to obtain structured visual and audio streams and to expand rubric descriptors into fine-grained behavioral indicators. Four modular components (Section~\ref{sec:video_reasoner}) then produce the assessment via the Scorer policy
\begin{equation}
    (\hat{s}, r) \sim \pi_\theta(\cdot \mid V, c, R(c, \cdot), \mathcal{B}(c, \cdot), A),
    \label{eq:policy}
\end{equation}
where $\hat{s}$ is the predicted score, $r$ its rationale, and $A$ the verified evidence chain assembled by the pipeline (Section~\ref{sec:video_reasoner}). Only $\pi_\theta^*$ is trained offline by supervised fine-tuning followed by rubrics-based reinforcement learning (Section~\ref{sec:grounded_scoring}).

\subsection{Multimodal Preprocessing}
\label{sec:preprocessing}

Before entering the agent framework, PhoenixNest-Video transforms raw video interviews into structured multimodal inputs through three parallel processing streams.

\noindent \textbf{Visual Stream.} We uniformly sample 32 frames per video to capture the candidate's expressions, body language, and overall presentation demeanor, with 32 chosen empirically as the accuracy peak (Section~\ref{sec:deep_analysis}). We additionally extract slide images to capture presentation content. This global sample is criterion-agnostic and runs in parallel with the clip partitioning of the Grapher (Section~\ref{sec:video_reasoner}). Key moments arise downstream, per criterion, as the clips that survive rubric-conditioned retrieval and multimodal verification.

\noindent \textbf{Audio Stream.} We extract the audio track from each video, apply ZipEnhancer for noise reduction and speech clarity enhancement, and transcribe the enhanced audio using Whisper-Large-v3~\cite{radford2023robust}.

\noindent \textbf{Rubric Decomposition.} For each criterion $c \in C$ with rubric descriptors $R(c, s)$ for $s \in \{0, 1, 2\}$, we use an MLLM to expand the concise rubric text into fine-grained behavioral indicators:
\begin{equation}
    \mathcal{B}(c, s) = \{b_1, b_2, \dots, b_m\} \leftarrow \text{MLLM}(R(c, s)),
    \label{eq:behavior_gen}
\end{equation}
where $m$ is the number of behavioral indicators generated for the pair $(c, s)$, ranging from three to six in our implementation, and each $b_i$ describes a concrete behavioral manifestation corresponding to score level $s$. For instance, for Critical Thinking at score level $s=2$, the decomposition produces indicators such as ``analyzes the problem from multiple complementary angles'' and ``supports claims with specific evidence''; at $s=0$, it produces ``provides circular or unsubstantiated reasoning'' and ``fails to engage with the substance of the question.'' These behavioral indicators ground the reward signals during training and direct the retrieval queries during inference.

\subsection{Pipeline Components}
\label{sec:video_reasoner}

To ensure that every score decision is anchored in verifiable video content, PhoenixNest-Video passes each video through four sequential stages at inference time, transforming a long-form video interview into a structured, auditable assessment through the Grapher, Retriever, Verifier, and Scorer.

\noindent \textbf{Grapher.}
We partition the video $V$ at 1.0~FPS into clips $\{V_1, \dots, V_n\}$ of $K = 64$ frames each and prompt an MLLM to extract open-vocabulary semantic mentions $E_i$ from each clip together with its aligned transcript $C_i$. A mention is an entity (a person, object, or presentation material, stored as a short name with a description), an action, or a scene label marking the interview phase. The vocabulary is open and drawn from the clip content itself. Two mentions are compared by cosine similarity on L2-normalized \texttt{[CLS]} embeddings from \texttt{BAAI/bge-large-en-v1.5},
\begin{equation}
    \mathrm{sim}(t_a, t_b) = \frac{\mathbf{e}_a \cdot \mathbf{e}_b}{\|\mathbf{e}_a\|\,\|\mathbf{e}_b\|},
    \label{eq:entity_affinity}
\end{equation}
and are merged into one prototype entity when $\mathrm{sim} > \tau$. The semantic graph $\mathcal{G} = (\mathcal{V}, \mathcal{E})$ has clip nodes $\mathcal{V} = \{v_i\}_{i=1}^n$, and an edge joins two clips whenever they share a prototype entity. Edges therefore link temporally separated moments that carry semantically equivalent content, which is what a rubric criterion typically requires: a candidate may demonstrate critical thinking in the self-introduction, again in the project presentation, and again under questioning. $\mathcal{G}$ is built once and reused across all criteria.

\noindent \textbf{Retriever.}
For each criterion $c_i$, the Retriever (i) takes the behavioral indicators $\mathcal{B}(c_i, s)$ from Eq.~\ref{eq:behavior_gen} and extracts semantic keywords $\mathcal{K}$, (ii) prompts an MLLM to refine each indicator $b_j$ into targeted queries $Q(b_j)$, and (iii) matches them against $\mathcal{G}$ under Eq.~\ref{eq:entity_affinity}. A query keyword that matches a prototype entity above $\theta$ pulls in \emph{every} clip node sharing that prototype, so the merged-entity edges act as an inverted index that surfaces temporally distant evidence a clip-local match would miss. A second pass adds nodes whose own attributes exceed $\theta$ directly, and the union is re-ranked by average embedding similarity, keeping the top-$N_r$ candidate clips $\mathcal{R}$. $\mathcal{R}$ is topically relevant but may include false positives.

\noindent \textbf{Verifier.}
For each indicator $b_j$ and clip $v_i \in \mathcal{R}$, an MLLM is queried with the binary question ``Does this clip demonstrate $b_j$?'' across visual, audio, and textual modalities with equal weight; $v_i$ is retained in $\mathcal{R}'$ if at least one modality answers positively above confidence $\delta$. The retained clips form the reasoning chain $A$, a criterion-specific evidence package in which each entry records the clip identifier, its temporal range in the interview, the matched indicator $b_j$, and per-modality flags marking which channels supported the match. Clips on which no modality passes $\delta$ are dropped, and the surviving entries are ordered chronologically, so $A$ reads as a time-ordered evidence trail that a reviewer can replay against the recording.

\noindent \textbf{Scorer.}
The trained Scorer $\pi_\theta^*$ is invoked on the reasoning chain $A$, the rubric $R(c_i, \cdot)$, and the behavioral indicators $\mathcal{B}(c_i, \cdot)$ for all score levels, producing the predicted score $\hat{s}$, its rationale $r$, textual feedback $F$, and references to the candidate's materials drawn from $\mathcal{R}'$. Because $\pi_\theta^*$ has been optimized (Section~\ref{sec:grounded_scoring}) for rubric-faithful rationales and institutionally calibrated scores, its output is directly consumable as the final report.

\subsection{Scorer Training}
\label{sec:grounded_scoring}

The Scorer component $\pi_\theta^*$ invoked at the assessment stage is trained offline in two steps: supervised fine-tuning followed by rubrics-based reinforcement learning.

\noindent \textbf{Supervised Fine-Tuning.}
We fine-tune the base model on expert annotation pairs $\mathcal{D}_{\text{SFT}} = \{(V_j, c_j, s_j^*, r_j^*)\}$, establishing the basic evaluation format, output structure, and initial score distribution.

\noindent \textbf{Rubrics-based Reinforcement Learning.}
Recent clinical MLLM work shows that continuous rubric-based rewards can provide informative supervision when binary feedback is sparse~\cite{fan2026oralgpt}, and structured reward shaping has been used more broadly to densify the learning signal in RL post-training~\cite{shi2026spader}. A binary exact-match GRPO reward is likewise too sparse here: with 18 criteria on a three-level scale and a total spanning $[0, 36]$, exact matches are rare in early training and most rollouts receive zero gradient. We therefore design two complementary reward signals that decompose assessment quality along orthogonal axes.

\noindent \textbf{Alignment Reward.} The alignment reward $R_{\text{align}}$ measures whether the generated rationale correctly references and applies the rubric criteria for the predicted score level. We employ an independent LLM as a judge that receives the rubric descriptor $R(c, s)$, the behavioral indicators $\mathcal{B}(c, s)$, the model's predicted score $\hat{s}$, and the generated rationale $r$, and evaluates:
\begin{equation}
    R_{\text{align}} = \text{LLM}_{\text{judge}}(R(c, s),\; \mathcal{B}(c, s),\; \hat{s},\; r),
    \label{eq:r_align}
\end{equation}
where $\text{LLM}_{\text{judge}}$ assesses whether the rationale faithfully grounds its reasoning in the rubric descriptors and whether the cited behavioral evidence supports the assigned score level.

\noindent \textbf{Differentiation Reward.} The differentiation reward $R_{\text{diff}}$ measures the agreement between the model's predicted total score and the expert-assigned total score at the institutional grading level. Let $L(\cdot)$ denote the level mapping function that assigns a total score to one of four institutional levels: $[0, 18)$, $[18, 24)$, $[24, 30)$, $[30, 36]$. The reward is:
\begin{equation}
    R_{\text{diff}} = \begin{cases}
        1.0 & \text{if } L(\hat{S}_{total}) = L(S^*_{total}), \\
        0.5 & \text{if } |L(\hat{S}_{total}) - L(S^*_{total})| = 1, \\
        0.0 & \text{if } |L(\hat{S}_{total}) - L(S^*_{total})| \geq 2,
    \end{cases}
    \label{eq:r_diff}
\end{equation}
where $i$ indexes the rubric criteria, $\hat{S}_{total} = \sum_{i=1}^{N} \hat{s}_i$ is the model's predicted total score summed over the $N$ criterion-level predictions $\hat{s}_i$, and $S^*_{total}$ is the expert total score.

The total reward is $R = \lambda_1 R_{\text{align}} + \lambda_2 R_{\text{diff}}$, and we optimize $\pi_\theta$ with standard GRPO~\cite{shao2024deepseekmath} using group-normalized advantages and a KL penalty $\beta\,\mathrm{KL}(\pi_\theta\|\pi_{\text{ref}})$ against the reference policy.

\begin{table*}[t]
    \centering
    \footnotesize
    \setlength{\tabcolsep}{4pt}
    \renewcommand{\arraystretch}{1.05}
    \resizebox{\textwidth}{!}{%
    \begin{tabular}{llcccccc}
        \toprule
        & & & \multicolumn{3}{c}{\textbf{Total Score}} & \multicolumn{2}{c}{\textbf{Per-Criterion}} \\
        \cmidrule(lr){4-6} \cmidrule(lr){7-8}
        Category & Model & Params & ACC $\uparrow$ & MAE $\downarrow$ & W Dist $\downarrow$ & QWK $\uparrow$ & MAE $\downarrow$ \\
        \midrule
        \multirow{5}{*}{\textit{Proprietary}}
        & GPT-5.4~\cite{singh2025openai} & - & 0.4300 & 10.7350 & 10.7283 & 0.0618 & 0.6350 \\
        & Claude-opus-4-6~\cite{claude4_5} & - & 0.4450 & 12.9750 & 12.9583 & 0.0915 & 0.7456 \\
        & Gemini-3.1-pro-preview~\cite{team2023gemini} & - & 0.8500 & 5.4633 & 4.2333 & \textbf{0.1733} & 0.5092 \\
        & Grok-4.1~\cite{grok_4} & - & \underline{0.9000} & \underline{4.3517} & \underline{2.4733} & 0.1069 & 0.4797 \\
        \midrule
        \multirow{6}{*}{\textit{Open-Source}}
        & Qwen3.5-397B-A17B~\cite{bai2025qwen3} & 397B & 0.7800 & 6.8767 & 6.2833 & 0.1172 & 0.4947 \\
        & Qwen3.5-27B~\cite{bai2025qwen3} & 27B & 0.7800 & 6.9583 & 6.4883 & 0.1287 & 0.5267 \\
        & Kimi-K2.5~\cite{team2026kimi} & 1000B & 0.8050 & 6.9217 & 6.4783 & 0.1111 & 0.5406 \\
        & GLM-4.5v~\cite{Hong2025GLM45VAG} & 106B & 0.7688 & 6.9899 & 3.5528 & 0.0738 & 0.5137 \\
        \midrule
        \multirow{5}{*}{\textit{Video-Specific}}
        & Video-R1~\cite{feng2025video} & 7B & 0.7310 & 6.6447 & 3.5854 & 0.0223 & 0.4929 \\
        & VideoChat-R1~\cite{li2025videochat} & 7B & 0.8000 & 5.9983 & 2.7133 & 0.0701 & 0.4333 \\
        & VideoRFT~\cite{wang2025videorft} & 7B & 0.6300 & 10.4867 & 9.1483 & -0.0123 & 0.5850 \\
        & VideoLLaMA2~\cite{cheng2024videollama} & 7B & 0.6782 & 7.0230 & 4.2337 & 0.0032 & 0.5160 \\
        & VideoLLaMA3~\cite{zhang2025videollama} & 7B & 0.8528 & 5.6210 & 3.6870 & -0.0179 & \underline{0.4123} \\
        \midrule
        \rowcolor{OursRow}
        \multirow{4}{*}{\textit{Ours}}
        & PhoenixNest-Video (trained Scorer) & 8B & \textbf{0.9150} & \textbf{4.1316} & \textbf{2.0328} & 0.1352 & 0.4419 \\
        \rowcolor{OursRow}
        & PhoenixNest-Video (Qwen3-397B) & 397B & 0.8700\imp{+0.09} & 4.8450\imp{-2.03} & 3.2850\imp{-3.00} & 0.1552\imp{+0.04} & 0.4175\imp{-0.08} \\
        \rowcolor{OursRow}
        & PhoenixNest-Video (Kimi-K2.5) & 1000B & 0.8400\imp{+0.04} & 5.1250\imp{-1.80} & 3.8650\imp{-2.61} & \underline{0.1689}\imp{+0.06} & 0.4219\imp{-0.12} \\
        \rowcolor{OursRow}
        & PhoenixNest-Video (GLM-4.5v) & 106B & 0.8950\imp{+0.13} & 4.4150\imp{-2.57} & 2.6350\imp{-0.92} & 0.1171\imp{+0.04} & \textbf{0.3869}\imp{-0.13} \\
        \bottomrule
    \end{tabular}%
    }
    \caption{Main results on the VInterview-2025 test set. \textbf{Bold} indicates the best result per column; \underline{underline} indicates the second best. ``Proprietary'' refers to closed-source general-purpose models accessed via commercial APIs; ``Open-Source'' refers to publicly available general-purpose open models; ``Video-Specific'' refers to models specifically designed or optimized for video understanding and reasoning. Total score metrics measure alignment at the institutional grading level; per-criterion metrics measure agreement on individual 0--2 rubric scores. Missing or non-parseable model outputs are excluded from scoring, so the number of scored interviews is at most 200 and varies across models. For the three backbone-amplifier rows in the \textit{Ours} block, the small green value in parentheses, e.g.\ {\color{ImpGreen}\scriptsize($\Delta$+0.09)}, denotes the absolute change of the wrapped configuration relative to the same backbone's direct-prompt counterpart in the \textit{Open-Source} block; the sign already encodes the direction of improvement with respect to the metric.}
    \label{tab:main_results}
\end{table*}

\section{Experiments}
\label{sec:exp}
\begin{table*}[t]
\scriptsize
\centering
\setlength{\tabcolsep}{5pt}
\renewcommand{\arraystretch}{1.15}
\begin{adjustbox}{width=\linewidth}
\begin{tabular}{l | cccc}
\toprule
    \textbf{Model} &
    \textbf{Spearman $\rho$} &
    \textbf{Kendall $\tau$-b} &
    \textbf{C-index} &
    \textbf{Pearson $r$} \\
\midrule
    GPT-5.4 ~\cite{singh2025openai} & 0.3096 & 0.2218 & 0.6067 & 0.2702 \\
    Claude Opus 4.6~\cite{claude4_5} & \underline{0.3832} & \underline{0.2763} & \underline{0.6328} & \underline{0.3756} \\
    Gemini-3.1-Pro-Preview~\cite{team2023gemini} & 0.1775 & 0.1303 & 0.5571 & 0.1560 \\
    Grok-4.1~\cite{grok_4}  & 0.2962 & 0.2075 & 0.6022 & 0.2437 \\
    Doubao-Seed-1.8-251228~\cite{guo2025seed1}  & 0.2985 & 0.2158 & 0.5979 & 0.2440 \\
    Qwen3.5-397B-A17B~\cite{bai2025qwen3}  & 0.2973 & 0.2120 & 0.6018 & 0.2559 \\
    GLM-4.5v~\cite{Hong2025GLM45VAG}  & 0.3313 & 0.2350 & 0.6141 & 0.2926 \\
    Kimi-K2.5~\cite{team2026kimi}  & 0.1914 & 0.1436 & 0.5591 & 0.1849 \\
\midrule
    PhoenixNest-Video & \textbf{0.4437} & \textbf{0.3159} & \textbf{0.6579} & \textbf{0.4234} \\
\bottomrule
\end{tabular}
\end{adjustbox}
\caption{Macro-averaged performance on the RecruitView dataset~\cite{gupta2025recruitview} across all 12 targets on the 317-sample user-grouped test split. \textbf{Bold} indicates the best result and \underline{underline} the second best in each column.
All baseline rows report our own evaluation of the corresponding model under the prompt-based protocol.
Missing or non-parseable API outputs are excluded from scoring.}
\label{tab:recruitview}
\end{table*}
We design experiments around four research questions that together evaluate whether PhoenixNest-Video meets the practical demands of an admissions-facing copilot:
\begin{itemize}[noitemsep, topsep=2pt, leftmargin=*]
    \item \textbf{RQ1} How does PhoenixNest-Video compare with proprietary, open-source, and video-specialized baselines on real-world video interview assessment, and does the framework lift existing MLLM backbones beyond their direct-prompt performance?
    \item \textbf{RQ2} What is the individual contribution of each key component, namely the alignment reward $R_{\text{align}}$, the differentiation reward $R_{\text{diff}}$, and the retrieval-and-verification pipeline formed by the Grapher, Retriever, and Verifier?
    \item \textbf{RQ3} Does the framework remain effective when retrained under a different annotation schema, i.e., on a benchmark whose targets are continuous personality and performance dimensions rather than rubric-grounded ordinal scores?
    \item \textbf{RQ4} Does PhoenixNest-Video exhibit any systematic bias along sensitive demographic or disciplinary axes, and how sensitive are its scores to the number of sampled frames per video?
\end{itemize}

\subsection{Experimental Setup}

\noindent \textbf{Datasets.}
We evaluate on two complementary benchmarks.
\textbf{VInterview-2025} is a self-collected dataset of 491 real-world graduate admissions interviews recorded during a live admissions cycle at the participating institution. The recordings were obtained passively from the existing admissions workflow and at no point fed back into the decision process, so the panel scores reflect a genuine high-stakes evaluation rather than an annotation task performed for our work; this yields high-quality rubric labels while keeping the data collection free of any decision-altering intervention. Each interview lasts approximately 15 minutes and is rated by a three-member faculty panel on 18 rubric criteria (0--2 scale). The rubric is the institution's standard admissions instrument, maintained by the admissions committee and in operational use across earlier cycles; it was not written or modified for this study. All three panelists are faculty with admissions experience who received institutional training on the criteria and scoring standards, and each submits a separate 18-criterion score vector. Following the institution's own aggregation rule, the total-score ground truth is the arithmetic mean of the three faculty totals, $S_{\mathrm{GT}} = \frac{1}{3}\sum_{r=1}^{3} S^{(r)}$, and the per-criterion reference is the arithmetic mean of the three faculty criterion scores. We use 100 interviews for supervised fine-tuning, 191 for rubrics-based reinforcement learning, and hold out 200 for testing; the two training stages draw on non-overlapping candidates, and the 200 test interviews come from a different admissions round than the training data, so the evaluation is cross-round rather than a random split within one batch. To protect candidate privacy, the corpus stays on institutional infrastructure and is not publicly released.
\textbf{RecruitView}~\cite{gupta2025recruitview} is a public benchmark of 2{,}011 job-interview videos with expert annotations along 12 regression targets covering overall personality, speaking skills, confidence, the Big Five personality traits, and additional interview-performance dimensions, with the official user-stratified split of 1{,}404 training, 290 validation, and 317 test clips, on which we retrain and evaluate our framework.
\noindent \textbf{Baselines.}
We compare against three categories of MLLMs under the same prompt-based protocol: four proprietary~\cite{singh2025openai, claude4_5, team2023gemini, grok_4}, four open-source~\cite{bai2025qwen3, team2026kimi, Hong2025GLM45VAG}, and five video-specialized~\cite{feng2025video, li2025videochat, wang2025videorft, cheng2024videollama, zhang2025videollama}.

\noindent \textbf{Implementation Details.}
We use Qwen3-VL-8B-Instruct as the backbone, trained for 3 epochs with a batch size of 16 and a learning rate of $2\times10^{-5}$. The LLM judge for $R_{\text{align}}$ is GPT-5.4. Reward weights are $\lambda_1 = \lambda_2 = 0.5$ with $\beta = 0.05$. The retrieval-and-verification pipeline uses the untrained base model for graph construction and retrieval, while the final assessment is generated by the trained $\pi_\theta$. We use VLMEvalKit~\cite{duan2024vlmevalkit} to execute model inference and standardize output parsing during evaluation. Video processing uniformly samples 32 frames per video, while the Grapher partitions the video at 1.0~FPS into $K = 64$-frame clips; entity matching uses \texttt{BAAI/bge-large-en-v1.5} embeddings with merging threshold $\tau = 0.7$ and retrieval threshold $\theta = 0.5$. All experiments run on $4\times$A800 GPUs (80GB).

\noindent \textbf{Metrics.}
Following prior work~\cite{ke2024critiquellm, gu2024survey, li2024generation, gupta2025recruitview}, we report grade-level accuracy (ACC, computed over the institution's four-tier grading scheme, which aggregates raw rubric scores into the admissions-relevant grades faculty reviewers act on; a prediction counts as correct when its tier is within one tier of the expert tier), mean absolute error (MAE), Wasserstein distance (W~Dist) at the total-score level, and Quadratic Weighted Kappa (QWK) at the per-criterion level for VInterview-2025 test set.
For RecruitView~\cite{gupta2025recruitview}, we report Spearman $\rho$, Kendall $\tau$-b, C-index, and Pearson $r$, macro-averaged over all 12 targets following the evaluation protocol of the original paper.

\subsection{Main Results (RQ1)}

\noindent \textbf{Comparison with baselines.}
Table~\ref{tab:main_results} reports results on the VInterview-2025 test set. PhoenixNest-Video achieves the best total-score alignment and remains competitive on per-criterion agreement, despite using only an 8B backbone. Among the baselines, the two strongest proprietary models, Gemini-3.1-pro-preview and Grok-4.1, lead every other baseline on grade-level accuracy, while GPT-5.4 and Claude-opus-4-6 fall below the open-source and video-specialized groups; video-specialized models tend to be stronger on fine-grained per-criterion MAE. The fact that our 8B configuration matches or surpasses substantially larger proprietary and open-source models indicates that rubric-grounded training and evidence-grounded retrieval are more effective than raw scale for structured assessment.

\noindent \textbf{Framework as a backbone amplifier.}
The remaining \textit{Ours} rows wrap three large MLLMs (Qwen3.5-397B-A17B, Kimi-K2.5, GLM-4.5v) inside the PhoenixNest-Video framework. Each configuration consistently improves over the same backbone's direct-prompt counterpart on every metric, as shown by the green deltas next to each value. The largest absolute gain appears for GLM-4.5v, whose grade-level accuracy increases by more than ten points and whose per-criterion MAE drops by a comparable margin; Qwen3.5-397B-A17B and Kimi-K2.5 show similar trends. This confirms that the improvements stem from the framework itself rather than from a particular choice of backbone, supporting the claim that PhoenixNest-Video acts as a backbone-agnostic capability amplifier for existing MLLMs.

\subsection{Ablation Studies (RQ2)}
\begin{figure}[t]
    \centering
    \includegraphics[width=\linewidth]{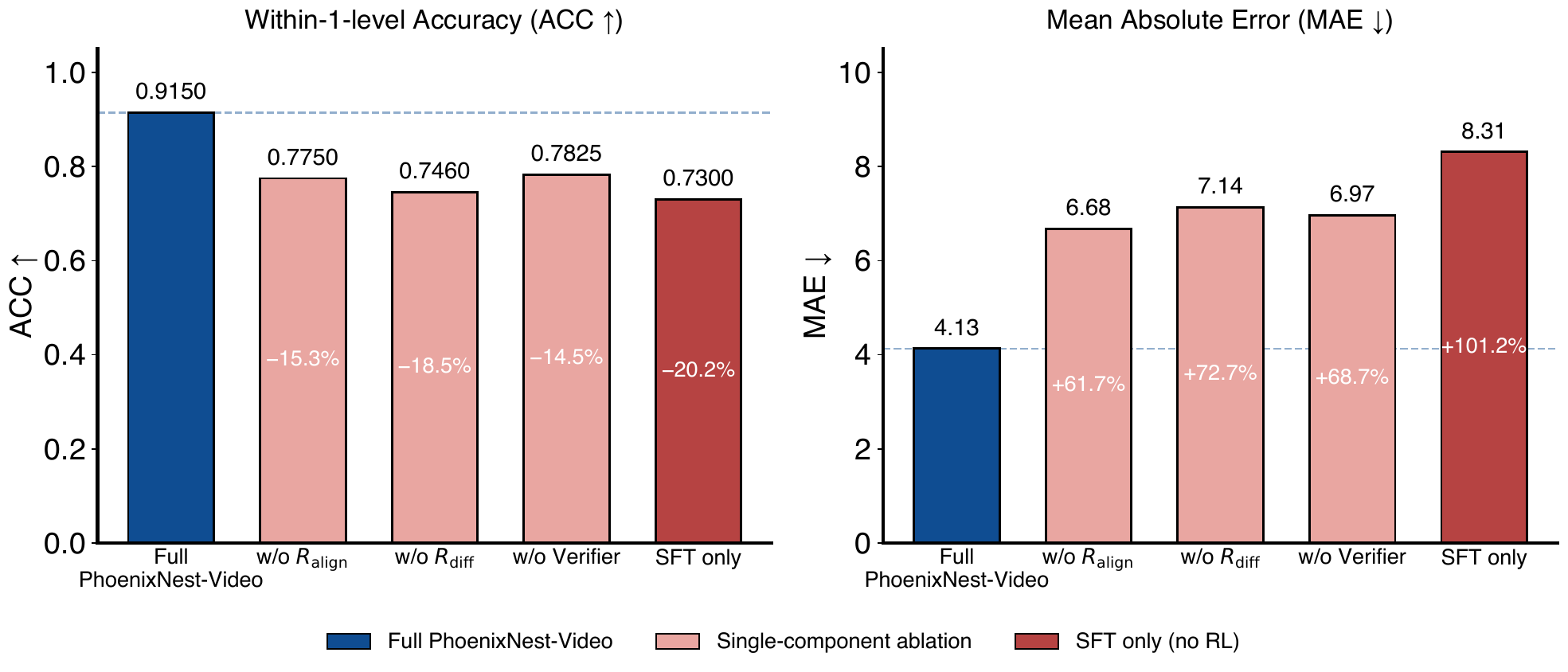}
    \caption{Ablation of PhoenixNest-Video on VInterview-2025. Removing any single component or skipping reinforcement learning consistently degrades grade-level accuracy and inflates MAE, indicating that the rubrics-based rewards and the retrieval-and-verification pipeline are complementary.}
    \label{fig:ablation}
\end{figure}
Figure~\ref{fig:ablation} isolates the contribution of each component by removing one at a time from the full framework. Removing either reward signal causes a substantial drop in accuracy and a sharp rise in MAE, with the differentiation reward $R_{\text{diff}}$ being slightly more impactful than the alignment reward $R_{\text{align}}$. This pattern indicates that the two rewards are complementary along orthogonal axes: $R_{\text{align}}$ enforces rubric-faithful rationales while $R_{\text{diff}}$ shapes the harder institutional-level score differentiation. Removing the Verifier, so that every retrieved clip is passed to the Scorer without cross-modal checking, produces a comparable degradation, confirming that verification contributes on top of the reward shaping rather than being subsumed by it. Finally, the SFT-only baseline yields the largest overall regression, validating that supervised fine-tuning alone is insufficient and that reinforcement training with structured rewards is necessary to internalize the rubric.

\subsection{Results on RecruitView (RQ3)}
\label{sec:recruitview}
To verify that the framework is not tied to the self-collected rubric schema, we retrain and evaluate PhoenixNest-Video on the RecruitView dataset~\cite{gupta2025recruitview} following its native train/test protocol. RecruitView differs from VInterview-2025 in two fundamental ways, its targets are continuous regression scores rather than 0--2 ordinal rubric scores, and they cover the Big Five personality traits, an overall-personality index, and six interview-performance dimensions rather than presentation- and Q\&A-oriented criteria. We compare against the proprietary and open-source models of Table~\ref{tab:main_results}, plus Doubao-Seed-1.8~\cite{guo2025seed1}, under the same prompt-based protocol. As Table~\ref{tab:recruitview} shows, PhoenixNest-Video attains the best score across all four rank-correlation and concordance metrics, while every baseline remains at moderate correlation levels, underscoring the intrinsic difficulty of video interview assessment even for strong proprietary models. The consistent advantage indicates that the rubric-grounded training procedure and the retrieval-and-verification pipeline are not coupled to the specific labeling convention on which they were originally developed, and that they remain effective when the target structure changes.

\subsection{Deep Analysis (RQ4)}
\label{sec:deep_analysis}

Beyond aggregate accuracy, we further analyze two properties of PhoenixNest-Video on VInterview-2025. First, we audit predictions for systematic score gaps along two sensitive axes, candidate gender and academic discipline. Second, we examine the sensitivity of the framework to the number of sampled frames per video, which is the most consequential preprocessing hyperparameter in our pipeline.

\begin{table}[t]
\centering
\small
\setlength{\tabcolsep}{8pt}
\renewcommand{\arraystretch}{1.1}
\begin{tabular}{lc}
\toprule
\textbf{Gender} & \textbf{Mean Total Score} \\
\midrule
Male & 25.86 \\
Female & 26.52 \\
\midrule
$\Delta$ (Female $-$ Male) & $+0.66$ ($1.8\%$) \\
\bottomrule
\end{tabular}
\caption{Mean predicted total score (max 36) by candidate gender on VInterview-2025. The aggregate gap is small relative to the score range and within-group variance.}
\label{tab:gender_bias}
\end{table}

\noindent \textbf{Gender Bias.}
Table~\ref{tab:gender_bias} reports the mean predicted total score on VInterview-2025 by candidate gender. The gap is small relative to the score range and the within-group spread, and the scores show no systematic preference for either gender. The Limitations section defines the scope of this audit.

\begin{figure}[t]
    \centering
    \includegraphics[width=\linewidth]{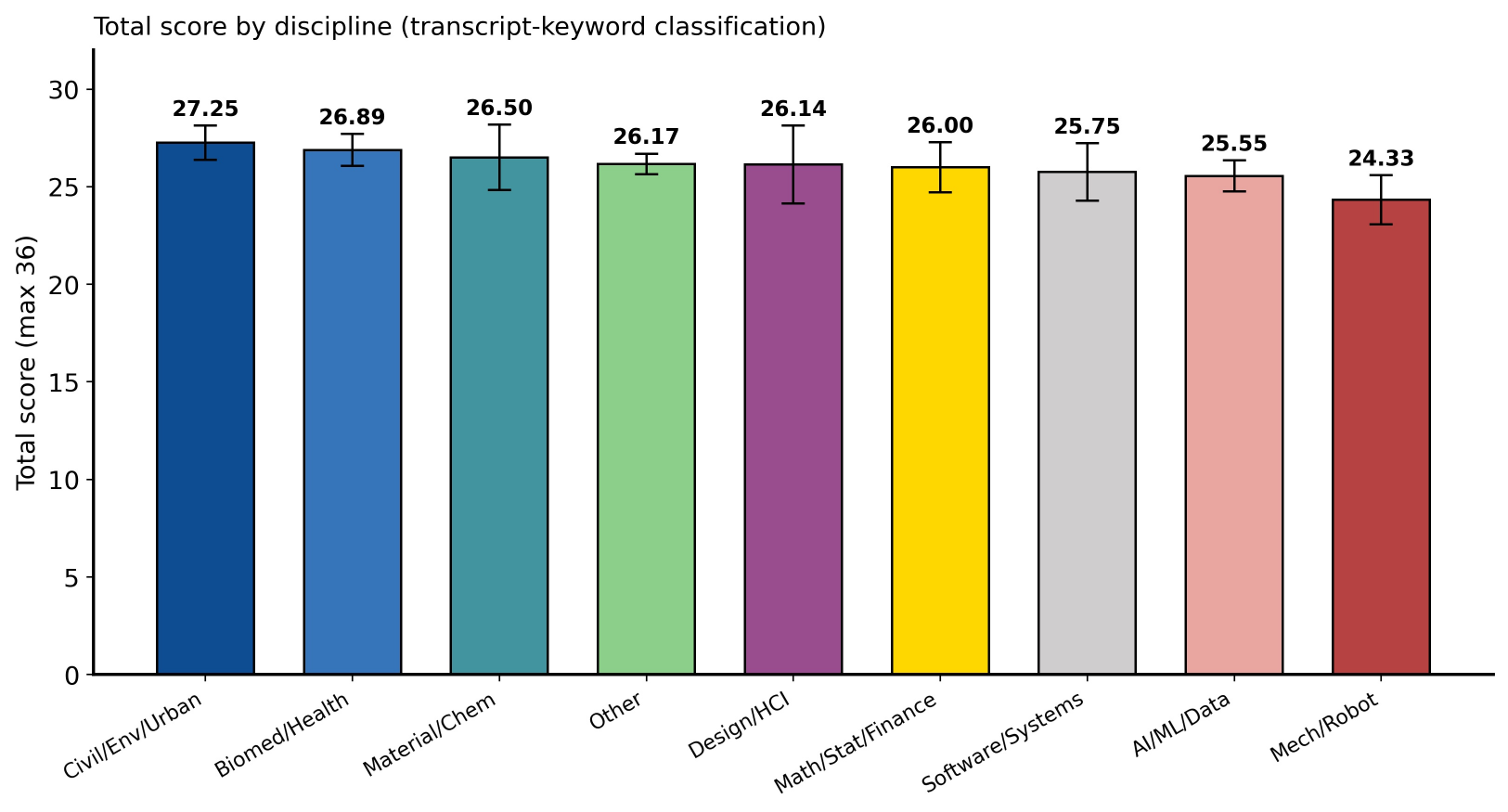}
    \caption{Mean predicted total score (max 36) by candidate discipline on VInterview-2025, where each candidate is assigned directly to their reported academic discipline. Error bars denote 95\% bootstrap confidence intervals.}
    \label{fig:subject_bias}
\end{figure}

\noindent \textbf{Subject Bias.}
Figure~\ref{fig:subject_bias} groups candidates by their academic discipline. The per-discipline means span a moderate range, with engineering-heavy disciplines slightly below the cross-discipline mean and verbal- or scenario-heavy disciplines slightly above. We attribute the residual spread to differences in evidence density: applicants in disciplines with richer verbal content tend to surface more rubric-aligned behavioral indicators during retrieval and verification. The overlap of per-discipline confidence intervals indicates that PhoenixNest-Video spreads its scores across fields rather than favoring a few.

\begin{table}[t]
\centering
\small
\setlength{\tabcolsep}{10pt}
\renewcommand{\arraystretch}{1.1}
\begin{tabular}{cc}
\toprule
\textbf{Sampled Frames} & \textbf{ACC} \\
\midrule
10 & 0.870 \\
16 & 0.895 \\
\rowcolor{OursRow}
32 & \textbf{0.915} \\
64 & 0.885 \\
\bottomrule
\end{tabular}
\caption{Effect of the number of uniformly sampled frames per video on grade-level accuracy on VInterview-2025. Accuracy peaks at 32 frames and then declines, indicating that denser sampling is not always beneficial.}
\label{tab:frame_ablation}
\end{table}

\noindent \textbf{Effect of Frame Sampling Density.}
Table~\ref{tab:frame_ablation} reports grade-level accuracy as we vary the number of uniformly sampled frames per video from 10 to 64. Accuracy rises sharply from 10 to 16 frames, plateaus through 32, and degrades at 64. Too few frames omit criterion-relevant evidence, whereas too many inflate the visual token budget and dilute attention across redundant near-duplicate frames. We therefore adopt 32 frames as the default throughout this paper.

\section{Conclusion}
\label{sec:conclusion}
We presented PhoenixNest-Video, a multimodal agent framework for automated video interview assessment. Rubrics-based Reinforcement Learning supplies dual rewards for rubric alignment and score-level differentiation, and a four-stage agent pipeline anchors each score to the elements of the candidate's materials that survive cross-modal verification. On VInterview-2025 and RecruitView, PhoenixNest-Video outperforms substantially larger proprietary and open-source baselines, showing that rubric grounding and evidence traceability matter more than raw scale for structured assessment.

\section*{Limitations}
Four conditions define where these results apply.

\noindent \textbf{Scope of the fairness audit.} The audit covers candidate gender and academic discipline in a single admissions cohort at one institution. Within these groups and this sample, predicted scores show no large aggregate difference, and the per-group sizes make only large disparities detectable. The admissions workflow the data comes from records no other candidate attributes, so ethnicity, nationality, accent, disability, socioeconomic background, and first-language background fall outside the audit; collecting them would require consent and data-protection provisions beyond the present ethical-review approval. The behavioral and linguistic signals the framework retrieves may also track culturally specific communication norms, or unequal access to interview coaching and recording conditions, rather than candidate capability.

\noindent \textbf{Language proficiency and transcript quality.} The interviews are conducted in English and many candidates are non-native speakers, which constrains the system in two ways. First, ASR accuracy degrades under strong accents, background noise, and low-quality recordings, and these errors propagate into the transcript stream the Grapher and the Retriever consume. Measuring this effect requires human reference transcripts, which VInterview-2025 does not include. Second, disfluent or error-prone speech can depress criteria that target content rather than language, so a candidate with limited fluency may lose points for a reason the rubric does not intend to measure. The Scorer is trained on faculty ratings and reproduces this confound. Separating language proficiency from the competencies the rubric targets calls for interview data with wider variation in recording conditions and speaker language background.


\noindent \textbf{Fixed design choices and deployment scope.} Two components are set once and not ablated. The entries of the reasoning chain $A$ reach the Scorer in chronological order, and LLMs are sensitive to the ordering of structured inputs~\cite{he2026order}, so an alternative ordering such as by verification confidence may shift the resulting scores. The pipeline also assumes that the full interview is available before assessment begins; extending it to streaming video, so that intermediate scores and evidence anchors appear while the interview unfolds, remains open. VInterview-2025 itself stays closed. The recordings are graduate admissions interviews with identifiable candidates, and no anonymization we can apply to video and voice removes that identifiability, so we keep the corpus on institutional infrastructure and release neither the recordings nor the derived benchmark.

\bibliography{custom}

\end{document}